\documentclass[11pt,a4paper]{article}
\usepackage[utf8]{inputenc}
\usepackage{amsmath}
\usepackage{amsfonts}
\usepackage{amssymb}
\usepackage{graphicx}
\usepackage{arxiv}
\usepackage{hyperref}
\usepackage{booktabs}
\usepackage{subcaption}
\usepackage{float}
\usepackage{tikz}

\usetikzlibrary{arrows.meta, positioning, shapes.geometric, shadows.blur, calc}
\usepackage[ruled,vlined]{algorithm2e}

\title{CROSS-DOMAIN ADVERSARIAL AUGMENTATION: STABILIZING GANS FOR MEDICAL AND HANDWRITING DATA SCARCITY}

\author{
  \textbf{Md. Sohanuzzaman Soad} \\
  Department of Computer Science and Engineering \\
  University of Asia Pacific \\
  Dhaka, Bangladesh \\
  \texttt{18101064@uap-bd.edu} \\
  \And
  \textbf{Mahady Al Hady} \\
  Department of Computer Science and Engineering \\
  University of Asia Pacific \\
  Dhaka, Bangladesh \\
  \texttt{18101072@uap-bd.edu} \\
  \AND
  \textbf{S M Rafiuddin} \\
  Department of Computer Science and Engineering \\
  University of Asia Pacific \\
  Dhaka, Bangladesh \\
  \texttt{rifat.cse@uap-bd.edu} \\
  \And
  \textbf{Sudip Ghose} \\
  Department of Computer Science and Engineering \\
  University of Asia Pacific \\
  Dhaka, Bangladesh \\
  \texttt{18101094@uap-bd.edu}
}

\date{}

\begin{document}

\maketitle

\begin{abstract}
Generative Adversarial Networks (GANs) offer a pragmatic route to mitigate data scarcity in vision tasks. We study generative augmentation across two low-resource domains: Bangla handwritten characters and chest X-ray imaging using DCGAN-style models trained at 64×64 resolution. We evaluate fidelity and diversity via Inception Score (IS), Fréchet Inception Distance (FID), and embedding visualizations (t-SNE/UMAP), and assess downstream utility by training classifiers on real versus real synthetic data. Our experiments show that generative augmentation improves sample diversity and yields consistent gains in classifier performance under limited-data regimes. We analyze stability enhancements (e.g., gradient-penalized objectives and spectral normalization) and report ablations on synthetic-to-real ratios and sample filtering. We discuss evaluation caveats for medical images, dataset licensing, and privacy risks associated with synthetic data. The resulting protocol is simple to reproduce and provides a strong baseline for applying generative augmentation to resource-constrained imaging tasks.
\end{abstract}

\vspace{0.5cm}

\section{Introduction}

\textbf{\textit{Deep learning}} has transformed numerous fields, from \textit{automation} and \textit{education} to \textit{healthcare} and \textit{finance}, but its progress depends critically on the availability of \textbf{large, well-labeled datasets}. In practice, many important domains remain \textbf{\textit{data-scarce}}: handwritten character recognition in low-resource languages such as \textbf{\textit{Bangla}} suffers from limited labeled corpora, while \textbf{\textit{medical imaging datasets}} are often restricted by privacy, collection cost, and ethical constraints. Conventional augmentation techniques, like \texttt{flipping}, \texttt{rotation}, or \texttt{contrast adjustments}, can't capture the root cause of real-world data distributions and only offer a limited amount of diversity. 

\textbf{\textit{Generative Adversarial Networks}} (\texttt{GANs}) have become a strong option for addressing this lack of data. By recognizing how the data is spread out, \texttt{GANs} can make realistic samples that add variety to training sets. Researchers have investigated using them for \textit{handwriting recognition}, \textit{natural images}, and more in clinical fields like \textbf{\textit{chest X-ray analysis}}. But there are still problems to solve, even with this promise. Training a \texttt{GAN} is known to be very unstable, and there is a chance of \textbf{\textit{mode collapse}} or \textbf{\textit{artifacts}} that make samples less useful. This study also has evaluated common metrics like \texttt{Inception Score} and \texttt{Fréchet Inception Distance} only give partial information, and their dependability varies depending on the context, especially when evaluating medical images using \texttt{ImageNet}-trained features. 

In this study, we deeply investigate \textbf{\textit{GAN-based augmentation}} in two different domains: \textbf{\textit{chest X-ray imaging}} and \textbf{\textit{handwritten characters in Bangla}}. Besides simply assessing the level of quality and diversity of the generated images, we intend to do more. Additionally, we intend to determine if the inclusion of \textbf{\textit{synthetic data}} consistently improves classification in environments with limited resources. In summary, this work offers \textbf{four main contributions}. First, it provides a \textbf{\textit{cross-domain evaluation}} of generative augmentation using a single framework applied to \textit{Bangla handwritten characters} and \textit{chest X-ray images}, two areas where labeled data remain limited. Second, it evaluates the \textbf{\textit{fidelity}} and \textbf{\textit{diversity}} of generated samples through metrics such as \texttt{IS}, \texttt{FID}, \texttt{KID}, and \texttt{precision, recall} scores, supported by \texttt{t-SNE} and \texttt{UMAP} visualizations. Third, it examines the effect of generative augmentation on downstream performance by comparing classifiers trained on real data alone, with common augmentations like \texttt{Mixup}, \texttt{Cutout}, \texttt{CutMix}, and \texttt{AugMix}, and with \texttt{GAN}-generated samples, showing clear gains under \textbf{\textit{low-data conditions}}. Finally, it introduces a \textbf{\textit{reproducible protocol}} that incorporates stable training methods such as \texttt{WGAN-GP} and \texttt{spectral normalization}, dataset-specific preprocessing, and ethical guidelines related to \textit{privacy} and \textit{licensing} in medical imaging. 

Lastly, we proposed a \textbf{\textit{recurrent approach}} that includes reliable training methods like \texttt{WGAN-GP} and \texttt{spectral normalization}, dataset-specific preparation strategies, and ethical guidelines for medical image privacy and licensing. 

\section{Literature Review}

Between \texttt{2018} and \texttt{2025}, studies on \textbf{\textit{Generative Adversarial Networks}} (\texttt{GANs}) advanced steadily in both building models and assessment. Architectures like \texttt{DCGAN} \cite{goodfellow2014generative} provided the initial momentum, which subsequently gave way to more stable formulations such \texttt{WGAN-GP} and the application of \textbf{\textit{spectral normalization}}. To improve \textbf{\textit{generalization}} and reduce \textbf{\textit{overfitting}}, researchers have investigated mixed techniques that link \texttt{GANs} to external classifiers during the past few years (\texttt{2022–2024}). As a demonstration, examine \textbf{\textit{Haque's (2021) External Classifier GAN}} (\texttt{EC-GAN}), in which the generator is directed by an auxiliary classifier instead of the discriminator. Compared to conventional augmentation or regularization methods, this configuration resulted in greater classification accuracy \cite{lim2024future, tripathi2022recent}. 

\textbf{\textit{Evaluation methods}} have developed in parallel with architecture. \textbf{\textit{Costa et al. (2021)}} tracked generator–discriminator evolution using \texttt{t-SNE} visualizations, which went beyond the early dependence on adversarial loss curves and provided understandable insights into convergence. While recent work suggests \texttt{Kernel Inception Distance} (\texttt{KID}) and \texttt{Precision–Recall} curves for fuller \textbf{\textit{fidelity–diversity assessment}}, standard quantitative metrics like \texttt{Inception Score} (\texttt{IS}) and \texttt{Fréchet Inception Distance} (\texttt{FID}) are still frequently used \cite{kucharski2025towards}. 

\textbf{\textit{Handwriting recognition}} and \textbf{\textit{medical imaging}} have both made extensive use of \texttt{GANs}. \textbf{\textit{Bowles et al. (2018)}} showed that \texttt{GAN}-generated \texttt{MR} images in medical imaging enhanced segmentation in \textbf{\textit{low-data circumstances}}; subsequent research up until \texttt{2023} showed higher \texttt{Dice} scores and better handling of class imbalance. The \texttt{BanglaLekha-Isolated} dataset initially appeared in handwriting research by \textbf{\textit{Biswas et al. (2017)}}. It was then augmented with \textbf{\textit{multimodal characteristics}} and \texttt{GAN}-based augmentation to enhance \textbf{\textit{Indic script recognition}}. \textbf{\textit{Sikder (2020)}} further advanced Bangla character recognition by combining \texttt{GAN}-generated samples with deeper \texttt{CNNs}, achieving validation accuracy above \texttt{99.5 percent} and demonstrating the strong benefit of \textbf{\textit{synthetic data}} with modern architectures \cite{xun2022generative, hassan2023smart}. 

Finally, recent studies highlight \textbf{\textit{interpretability}} and \textbf{\textit{stability}} as ongoing challenges. Approaches such as \textbf{\textit{progressive growing GANs}}, \textbf{\textit{attention-based discriminators}}, and novel loss functions have improved fidelity and training reliability, while expanding applications to \textbf{\textit{video synthesis}}, \textbf{\textit{speech generation}}, and \textbf{\textit{multimodal learning}}. Emerging work even explores \texttt{GANs} for artistic content creation; for instance, the \texttt{2025 arXiv} study \textbf{\textit{Generative Adversarial Networks Bridging Art and Machine}} describes how modern \texttt{GANs} now span both \textbf{\textit{photorealistic synthesis}} and \textbf{\textit{creative design}} \cite{efatinasab2025towards}. 

Overall, the literature reflects a \textbf{\textit{multifaceted trajectory}}: continuous architectural innovation, evolving evaluation protocols, and expanding application domains from \textbf{\textit{low-resource language processing}} to \textbf{\textit{clinical imaging}} yet \textbf{\textit{cross-domain studies}} that jointly evaluate handwriting and medical tasks remain limited, motivating the present work.

\section{Methodology}

\begin{figure}[!htbp]
\centering
\begin{tikzpicture}[
    font=\small,
    >=Stealth,
    node distance=1.5cm,
    arrow/.style={->, thick},
    block/.style={
        rectangle,
        rounded corners=4pt,
        draw=black,
        thick,
        minimum width=3.0cm,
        minimum height=1.0cm,
        align=center
    },
    dataset/.style={block, fill=green!35},
    gan/.style={block, fill=yellow!45},
    gen/.style={block, fill=red!30},
    metric/.style={block, fill=blue!25},
    classifier/.style={block, fill=cyan!35},
    pred/.style={block, fill=green!30},
    decision/.style={
        diamond,
        draw=black,
        thick,
        fill=gray!25,
        aspect=1.5,
        align=center,
        inner sep=2pt
    },
    note/.style={align=center, font=\small}
]

\node[decision] (pass) at (0,0) {Pass?};

\node[dataset] (data) at (4,2) {Dataset\\Images};

\node[gan] (gan) at (4,0) {Generative\\Adversarial\\Network};

\node[gen] (generated) at (4,-2.2) {Generated\\Images};

\node[metric] (metrics) at (0,-2.2) {Quality\\Metrics};

\node[classifier] (clf) at (8.5,0) {External\\Classifier};

\node[pred] (predicted) at (8.5,-2.2) {Predicted\\Class};

\node[
    rectangle,
    draw=black,
    thick,
    fill=black,
    minimum width=1.4cm,
    minimum height=1.4cm
] (sample) at (12,0) {};

\node[note, below=0.25cm of sample] 
{Image from\\validation data\\passed to classifier};

\draw[white, line width=4pt, rounded corners=6pt]
    ($(sample.center)+(-0.35,0.35)$)
    .. controls ($(sample.center)+(-0.55,-0.20)$) 
    and ($(sample.center)+(0.10,-0.55)$)
    .. ($(sample.center)+(0.42,-0.18)$);

\draw[white, line width=4pt, rounded corners=6pt]
    ($(sample.center)+(0.40,-0.18)$)
    .. controls ($(sample.center)+(0.30,0.15)$)
    and ($(sample.center)+(0.55,0.35)$)
    .. ($(sample.center)+(0.15,0.45)$);

\draw[arrow] (data) -- node[right, align=center] {Used to train GAN} (gan);

\draw[arrow] (gan) -- node[right, align=center] {Generate images\\of each class} (generated);

\draw[arrow] (generated.west) -- (metrics.east);

\draw[arrow] (metrics.north) -- node[left, align=center] {Evaluate the\\sample quality} (pass.south);

\draw[arrow] (pass.north) |- node[above, align=center, pos=0.75] {Pass quality metrics} (data.west);

\draw[arrow] (pass.east) -- node[above, align=center] {Fail quality metrics} (gan.west);

\draw[arrow] (data.east) -- ++(1.6,0) |- 
node[right, align=center, pos=0.25] {Use generated with\\original images to train} (clf.north);

\draw[arrow] (sample.west) -- (clf.east);

\draw[arrow] (clf) -- (predicted);

\node[note, above=0.15cm of data] {\textbf{Real Training Data}};
\node[note, below=0.15cm of generated] {\textbf{Synthetic Samples}};
\node[note, below=0.15cm of predicted] {\textbf{Final Output}};

\end{tikzpicture}

\caption{GAN-based augmentation and classification workflow. Dataset images are first used to train the GAN, which generates class-specific synthetic images. The generated samples are evaluated through quality metrics; failed samples are used to further refine the GAN, while accepted samples are combined with the original data to train an external classifier. Validation images are then passed to the trained classifier to obtain the predicted class.}
\label{fig:gan_pipeline_tikz}
\end{figure}
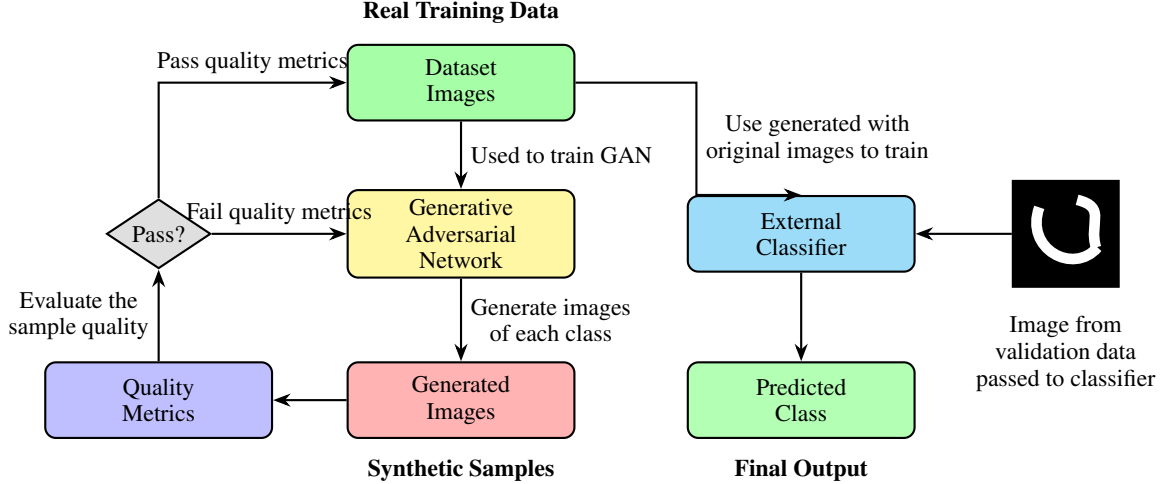

Our proposed framework follows a five-stage pipeline: (1) dataset selection and preprocessing, (2) generative model design, (3) stability-enhanced training, (4) dimensionality reduction for embedding visualization, and (5) evaluation with quantitative and qualitative metrics. Figure \ref{fig:gan_pipeline_tikz} provides an overview of the system. 



\subsection{Dataset Description}

\begin{figure}[htbp]
	\centering

	\begin{subfigure}[t]{0.48\linewidth}
		\centering
		\includegraphics[width=\linewidth]{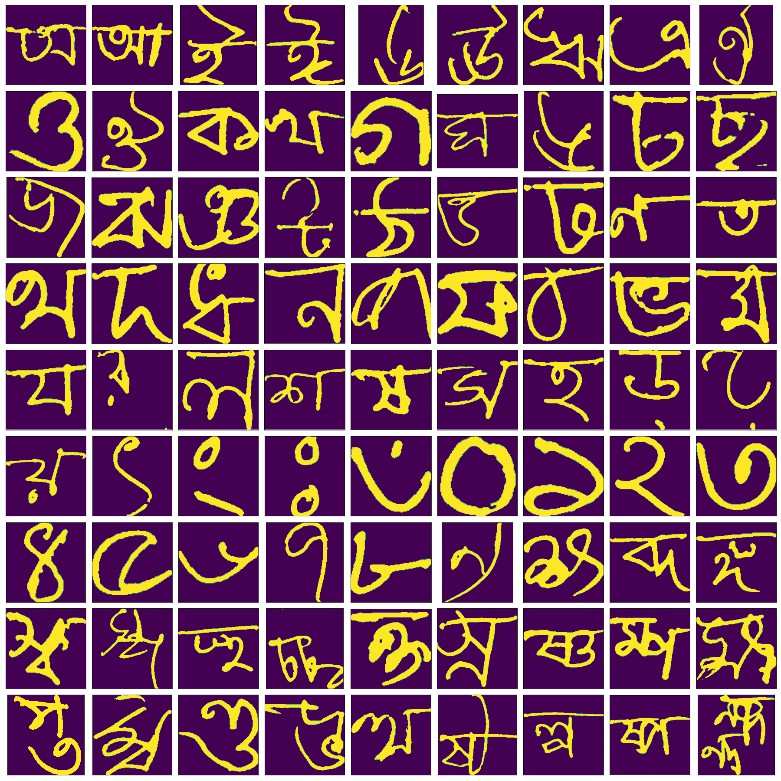}
		\caption{Sample images of the BanglaLekha Isolated Dataset.}
		\label{fig:banglalekha}
	\end{subfigure}
	\hfill
	\begin{subfigure}[t]{0.48\linewidth}
		\centering
		\includegraphics[width=\linewidth]{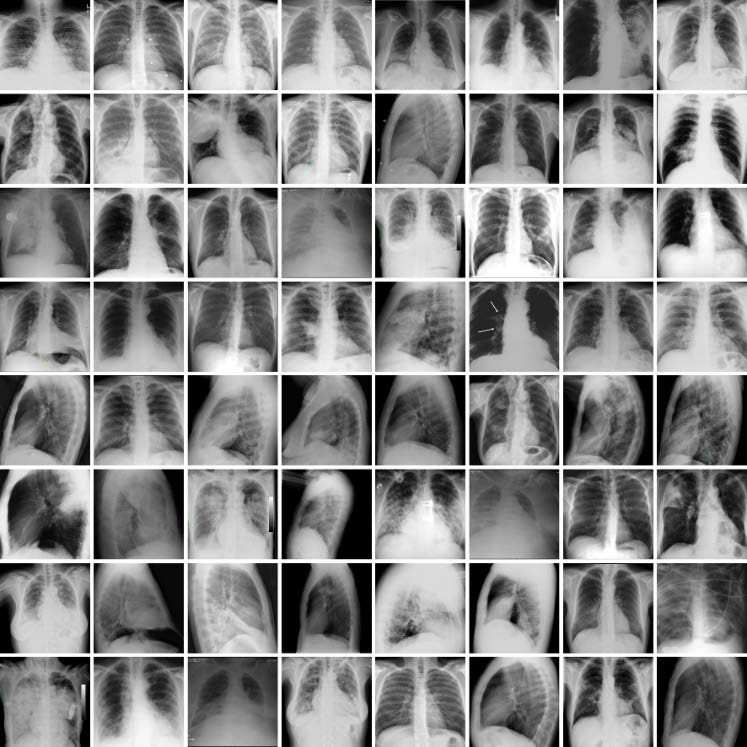}
		\caption{Sample images from the COVID-19 Chest X-Ray Dataset, illustrating variability across patient cases and imaging devices.}
		\label{fig:xray_samples}
	\end{subfigure}

	\caption{Representative sample images from the two datasets used in this study: Bangla handwritten characters and COVID-19 chest X-ray images.}
	\label{fig:dataset_samples}
\end{figure}

\texttt{BanglaLekha Isolated}, many handwritten \textbf{\textit{Bangla characters}} that comprise \textit{numerals}, \textit{simple letters}, and \textit{compound forms}, is the initial dataset used in this investigation. It includes over \texttt{160,000} samples taken from individuals in various parts of \textbf{\textit{Bangladesh}} and at various ages \cite{biswas2017banglalekha}. This dataset is particularly helpful for testing how effectively \texttt{GAN} models can replicate nuanced character shapes and maintain \textbf{\textit{fine-grained visual details}} due to the variety of \textbf{\textit{handwriting styles}}, as shown in Figure \ref{fig:banglalekha} of the thesis. 

Total 679 chest radiographs taken from patients across the world are included in the second dataset, the COVID-19 Chest X-ray dataset \cite{sekeroglu2020detection}. Due to differences in technology, patient anatomy, and health course, the photos show a variety of disorders, such as COVID-19, SARS, and MERS, and varied significantly in appearance. Sample images from the COVID-19 Chest X-Ray Dataset are shown in Figure \ref{fig:xray_samples}, which shows the minor, but important patterns required for a good diagnosis and shows diversity among patients and imaging equipment.

\subsection{Data Preprocessing}

\begin{figure}[!htbp]
	\centering

	\begin{subfigure}[t]{0.48\linewidth}
		\centering
		\includegraphics[width=\linewidth]{figures/figure_3.jpeg}
		\caption{Sample images from the COVID-19 Chest X-Ray Dataset, illustrating variability across patient cases and imaging devices.}
		\label{fig:xray_samples2}
	\end{subfigure}
	\hfill
	\begin{subfigure}[t]{0.48\linewidth}
		\centering
		\includegraphics[width=\linewidth]{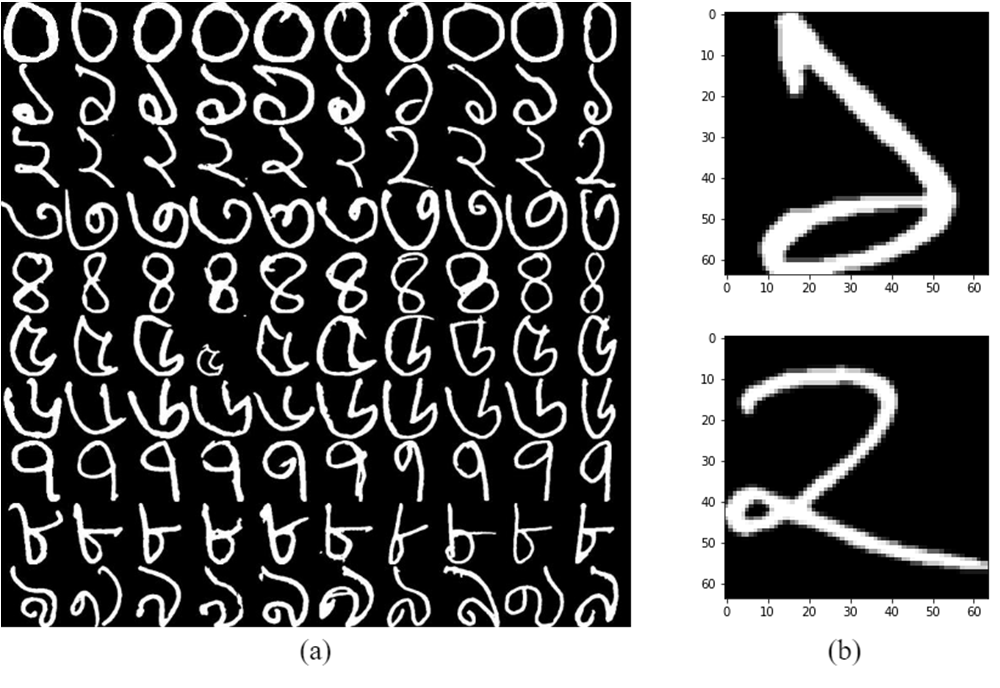}
		\caption{Example images from the preprocessed BanglaLekha Isolated dataset.}
		\label{fig:preprocessed}
	\end{subfigure}

	\caption{Representative sample images from the COVID-19 Chest X-Ray Dataset and the preprocessed BanglaLekha Isolated dataset.}
	\label{fig:xray_bangla_preprocessed}
\end{figure}

All photos were scaled to \texttt{64x64} pixels using \textbf{\textit{cubic interpolation}} to ensure consistent input for \texttt{GAN} training, which decreased computational complexity while maintaining crucial structural information. To provide constant gradient updates during training, pixel values were scaled to the interval \texttt{[-1, 1]}. Examples of the preprocessed photos after scaling and normalization are shown in Figure \ref{fig:preprocessed}. 

The image size was fixed at \texttt{64$\times$64} to balance \textbf{\textit{fidelity}} and \textbf{\textit{computational efficiency}}. Prior work demonstrated that \textbf{\textit{low-resolution GAN augmentation}} can still enhance classifier robustness in \textbf{\textit{data-scarce clinical contexts}}, while higher resolutions would exceed available \texttt{GPU} memory.

\subsection{GAN Architecture}

We implemented a \textbf{\textit{convolutional GAN}} inspired by \texttt{DCGAN} \cite{goodfellow2014generative}. The \textbf{\textit{generator}} transforms a \texttt{100}-dimensional latent vector into an image through stacked \textbf{\textit{transposed convolutional layers}}, each followed by \texttt{batch normalization} and \texttt{ReLU} activations, with a final \texttt{Tanh} output (Table \ref{tab:generator}). The \textbf{\textit{discriminator}} processes input images via \textbf{\textit{convolutional layers}} with \texttt{batch normalization} and \texttt{Leaky ReLU} activations, culminating in a \texttt{sigmoid} output for binary \texttt{real/fake} prediction (Table \ref{tab:discriminator}). 

\begin{table}[!htbp]
\centering
\caption{Generator and discriminator neural network architectures.}
\label{tab:gan_architectures}

\begin{subtable}[t]{0.48\linewidth}
\centering
\caption{Generative Neural Network Architecture}
\label{tab:generator}
\begin{tabular}{c c c}
\toprule
\textbf{Layer No.} & \textbf{Layer Type} & \textbf{Output Shape} \\
\midrule
1 & ConvTranspose2d & 512 x 4 x 4 \\
2 & BatchNorm2d & 512 x 4 x 4 \\
3 & ReLU & 512 x 4 x 4 \\
4 & ConvTranspose2d & 256 x 8 x 8 \\
5 & BatchNorm2d & 256 x 8 x 8 \\
6 & ReLU & 256 x 8 x 8 \\
7 & ConvTranspose2d & 128 x 16 x 16 \\
8 & BatchNorm2d & 128 x 16 x 16 \\
9 & ReLU & 128 x 16 x 16 \\
10 & ConvTranspose2d & 64 x 32 x 32 \\
11 & BatchNorm2d & 64 x 32 x 32 \\
12 & ReLU & 64 x 32 x 32 \\
13 & ConvTranspose2d & 3 x 64 x 64 \\
14 & Tanh & 3 x 64 x 64 \\
\bottomrule
\end{tabular}
\end{subtable}
\hfill
\begin{subtable}[t]{0.48\linewidth}
\centering
\caption{Discriminator Neural Network Architecture}
\label{tab:discriminator}
\begin{tabular}{c c c}
\toprule
\textbf{Layer No.} & \textbf{Layer Type} & \textbf{Output Shape} \\
\midrule
1 & Conv2d & 64 x 32 x 32 \\
2 & BatchNorm2d & 64 x 32 x 32 \\
3 & LeakyReLU & 64 x 32 x 32 \\
4 & Conv2d & 128 x 16 x 16 \\
5 & BatchNorm2d & 128 x 16 x 16 \\
6 & LeakyReLU & 128 x 16 x 16 \\
7 & Conv2d & 256 x 8 x 8 \\
8 & BatchNorm2d & 256 x 8 x 8 \\
9 & LeakyReLU & 256 x 8 x 8 \\
10 & Conv2d & 512 x 4 x 4 \\
11 & BatchNorm2d & 512 x 4 x 4 \\
12 & LeakyReLU & 512 x 4 x 4 \\
13 & Conv2d & 1 x 1 x 1 \\
14 & Flatten & 1 \\
15 & Sigmoid & 1 \\
\bottomrule
\end{tabular}
\end{subtable}

\end{table}

Conversely, the \textbf{\textit{discriminator network}} processes input images through \textbf{\textit{convolutional layers}} interleaved with \texttt{batch normalization} and \texttt{Leaky ReLU} activations. The network outputs a single probability indicating whether the input is \texttt{real} or \texttt{synthetic}. Table \ref{tab:discriminator} details the architecture.

\texttt{DCGAN} was specifically chosen due to its \textbf{\textit{interpretability}} and \textbf{\textit{training stability}} in \textbf{\textit{low-sample regimes}}. This decision allows for direct \textbf{\textit{cross-domain comparison}} and is in accordance with the study's objective of a \textbf{\textit{unified reproducible pipeline}} rather than architectural distinctness. 

\subsection{GAN Training Process}

\texttt{GAN} training took place across \texttt{300} epochs, using the \texttt{Adam} optimizer for both networks and a batch size of \texttt{32}. A local computer with an \texttt{NVIDIA GTX 1080 Ti GPU} was used for training, and \texttt{Google Colab Pro} was also used to boost \textbf{\textit{computational efficiency}}. The \texttt{BanglaLekha} dataset took about \texttt{2.5 hours} to train on \texttt{Colab}, while the \texttt{COVID-19} dataset took about \texttt{4 hours}. The training procedure follows the conventional \textbf{\textit{minimax game}}, in which the \textbf{\textit{discriminator}} constantly improves its capacity to identify \texttt{fake images} while the \textbf{\textit{generator}} tries to trick it. This \textbf{\textit{adversarial training loop}} is shown in Algorithm \ref{alg:gan_training}, which demonstrates how both networks continuously improve their performance. 

\begin{algorithm}[!htbp]
\caption{Minibatch stochastic gradient descent training of generative adversarial networks}
\label{alg:gan_training}

\KwIn{Number of training iterations, minibatch size $m$, discriminator steps $k$, noise prior $p_g(z)$, data distribution $p_{\text{data}}(x)$}
\KwOut{Trained generator $G$ and discriminator $D$}

\For{number of training iterations}{
    
    \For{$k$ steps}{
        Sample minibatch of $m$ noise samples 
        $\{z^{(1)}, \ldots, z^{(m)}\}$ from noise prior $p_g(z)$\;
        
        Sample minibatch of $m$ examples 
        $\{x^{(1)}, \ldots, x^{(m)}\}$ from data-generating distribution $p_{\text{data}}(x)$\;
        
        Update the discriminator by ascending its stochastic gradient:
        \[
        \nabla_{\theta_d}
        \frac{1}{m}
        \sum_{i=1}^{m}
        \left[
        \log D\left(x^{(i)}\right)
        +
        \log\left(1-D\left(G\left(z^{(i)}\right)\right)\right)
        \right]
        \]
    }
    
    Sample minibatch of $m$ noise samples 
    $\{z^{(1)}, \ldots, z^{(m)}\}$ from noise prior $p_g(z)$\;
    
    Update the generator by descending its stochastic gradient:
    \[
    \nabla_{\theta_g}
    \frac{1}{m}
    \sum_{i=1}^{m}
    \log\left(1-D\left(G\left(z^{(i)}\right)\right)\right)
    \]
}

\end{algorithm}

\subsection{Dimensionality Reduction and Visualization}

\textbf{\textit{Dimensionality reduction algorithms}} were used to measure the extent to which \textbf{\textit{synthetic images}} resemble the distribution of \textbf{\textit{real data}}. First, the feature space was reduced while preserving the directions of highest variance using \texttt{Principal Component Analysis} (\texttt{PCA}). \texttt{t-SNE}, which highlights local structure, cluster formation, and neighborhood interactions, was then used to map the reduced features into two dimensions. When combined, these procedures provide information about the total diversity of the samples produced as well as \textbf{\textit{class separability}}. The algorithmic flow of the \texttt{t-SNE} process used for visualization is shown in Algorithm \ref{alg:tsne_algorithm}. 

\begin{algorithm}[!htbp]
\caption{Simple version of t-Distributed Stochastic Neighbor Embedding}
\label{alg:tsne_algorithm}

\KwData{Data set $X=\{x_1,x_2,\ldots,x_n\}$; cost function parameter: perplexity $Perp$; optimization parameters: number of iterations $T$, learning rate $\eta$, momentum $\alpha(t)$}
\KwResult{Low-dimensional data representation $Y^{(T)}=\{y_1,y_2,\ldots,y_n\}$}

Compute pairwise affinities $p_{j|i}$ with perplexity $Perp$ using Equation~(1)\;

Set
\[
p_{ij}=\frac{p_{j|i}+p_{i|j}}{2n}
\]

Sample initial solution
\[
Y^{(0)}=\{y_1,y_2,\ldots,y_n\}
\]
from $\mathcal{N}(0,10^{-4}I)$\;

\For{$t=1$ \KwTo $T$}{
    Compute low-dimensional affinities $q_{ij}$ using Equation~(4)\;
    
    Compute gradient
    \[
    \frac{\delta C}{\delta Y}
    \]
    using Equation~(5)\;
    
    Set
    \[
    Y^{(t)}
    =
    Y^{(t-1)}
    +
    \eta \frac{\delta C}{\delta Y}
    +
    \alpha(t)\left(Y^{(t-1)}-Y^{(t-2)}\right)
    \]
}

\end{algorithm}

\section{Experiments and Results}

\subsection{Visual Assessment of Generated Samples}

\begin{figure}[htbp]
	\centering

	\begin{subfigure}[t]{0.48\linewidth}
		\centering
		\includegraphics[width=\linewidth]{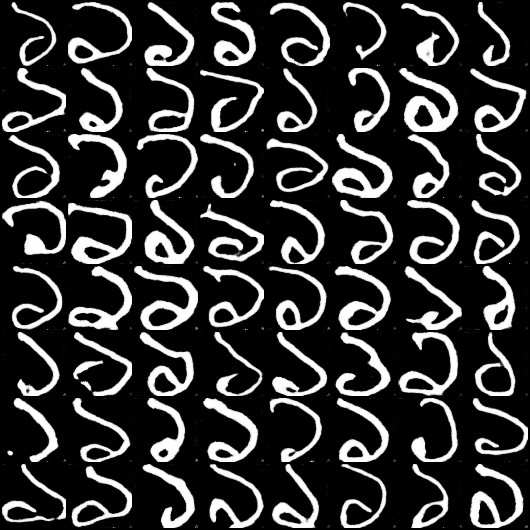}
		\caption{Images generated of Bangla Numeric Characters by GAN after 200 epochs of training.}
		\label{fig:generated_bangla}
	\end{subfigure}
	\hfill
	\begin{subfigure}[t]{0.48\linewidth}
		\centering
		\includegraphics[width=\linewidth]{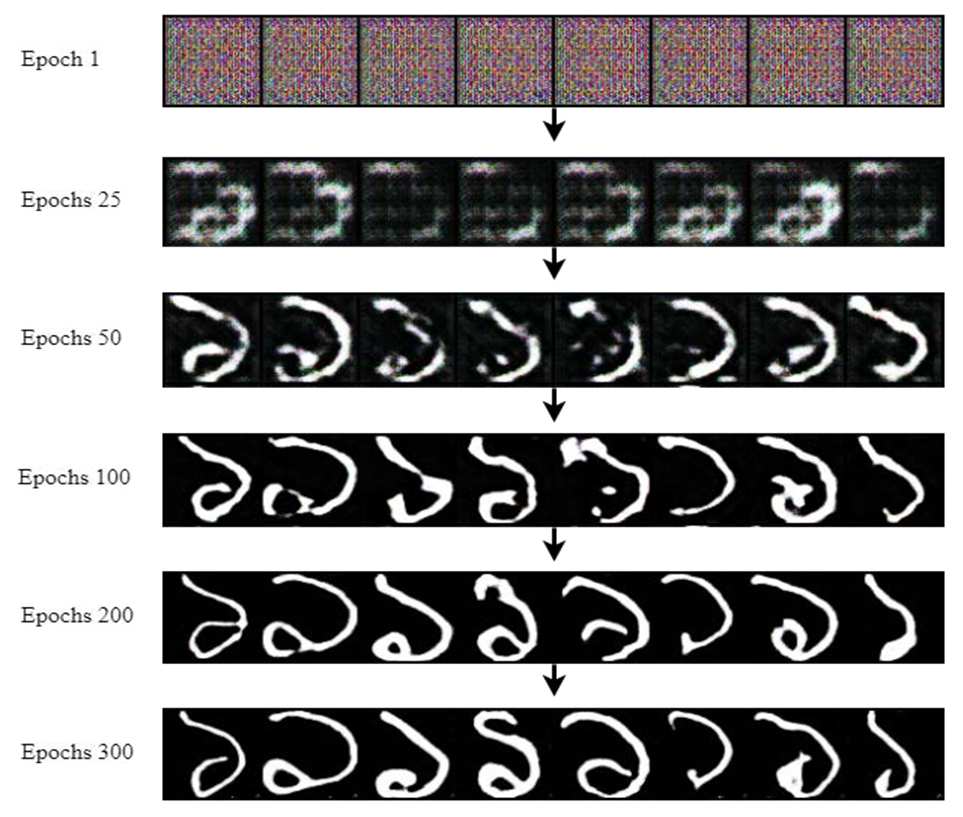}
		\caption{Generated Images of Bangla Numeric Characters by GAN at different stages of training.}
		\label{fig:generated_bangla_stages}
	\end{subfigure}

	\vspace{0.4cm}

	\begin{subfigure}[t]{0.48\linewidth}
		\centering
		\includegraphics[width=\linewidth]{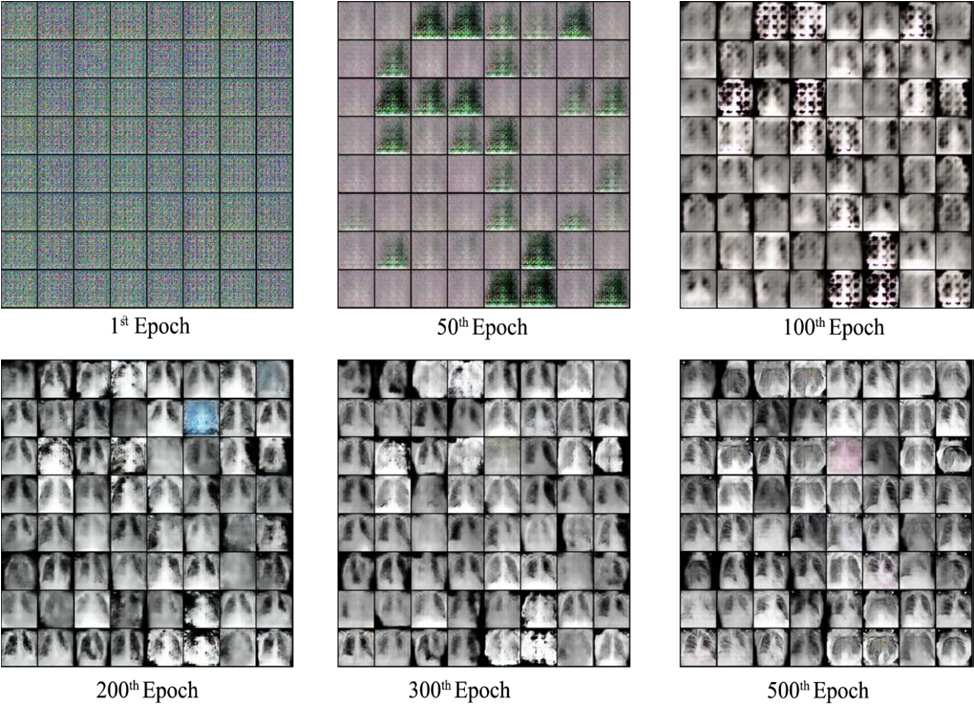}
		\caption{Generated Sample from COVID-19 Chest X-ray Dataset During Training.}
		\label{fig:xray_training}
	\end{subfigure}
	\hfill
	\begin{subfigure}[t]{0.48\linewidth}
		\centering
		\includegraphics[width=0.85\linewidth]{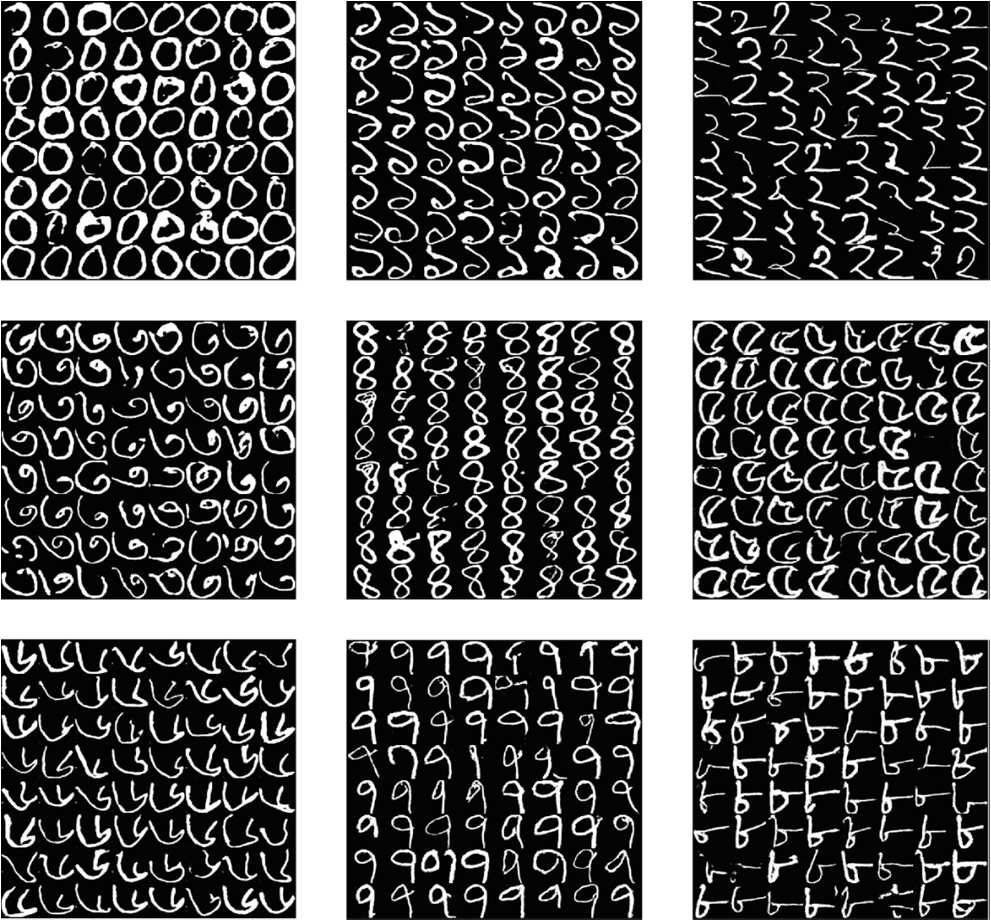}
		\caption{Final generated individual character of the Bangla Lekha Isolated dataset.}
		\label{fig:final_bangla}
	\end{subfigure}

	\caption{Visual assessment of GAN-generated samples across the BanglaLekha Isolated and COVID-19 Chest X-ray datasets.}
	\label{fig:generated_samples_grid}
\end{figure}

Over training epochs, the \textbf{\textit{generated images' quality}} significantly improved. Early \textbf{\textit{Bangla number}} examples were noisy and unstructured, shown in Figures \ref{fig:generated_bangla} and \ref{fig:generated_bangla_stages}. However, by the year \texttt{200}, the characters had \textbf{\textit{neat edges}}, \textbf{\textit{consistent strokes}}, and \textbf{\textit{stylistic diversity}} that reflected variations in handwriting. Medical data showed a similar pattern: The model created the \textbf{\textit{chest X-ray pictures}} shown in Figure \ref{fig:xray_training}, which show the general \textbf{\textit{anatomical structure}} of the rib cages and lungs. 

While \textbf{\textit{subtle artifacts}} remain, the \textbf{\textit{generated images}} demonstrate the model’s ability to reproduce \textbf{\textit{medically relevant patterns}}. Final samples of \textbf{\textit{Bangla characters}} in Figure \ref{fig:final_bangla} appear visually indistinguishable from \textbf{\textit{real samples}} to non-specialists, highlighting their potential for \textbf{\textit{data augmentation}}.
\subsection{Quantitative Evaluation with IS and FID}

\begin{table}[htbp]
\centering
\caption{Quantitative evaluation of generated image quality using Inception Score (IS) and Fréchet Inception Distance (FID) for the DCGAN model across the BanglaLekha Isolated and COVID-19 Chest X-ray datasets. Higher IS indicates greater sample diversity and recognizability, while lower FID indicates closer similarity between real and generated image distributions.}
\label{tab:is_fid}
\begin{tabular}{l l c c}
\toprule
\textbf{Model} & \textbf{Dataset} & \textbf{Inception} & \textbf{FID} \\
\textbf{Architecture} & & \textbf{Score} & \textbf{Score} \\
\midrule
DCGAN & BanglaLekha Isolated & 1.04 & 0.24 \\
DCGAN & COVID-19 Chest X-ray & 2.27 & 0.12 \\
\bottomrule
\end{tabular}
\end{table}

Table \ref{tab:is_fid} summarizes the \textbf{\textit{intrinsic evaluation metrics}}. For the \texttt{BanglaLekha-Isolated} dataset, the \texttt{Inception Score} reached \texttt{1.04} and the \texttt{Fréchet Inception Distance} was \texttt{0.24}, while the \texttt{COVID-19 chest X-ray} dataset achieved an \texttt{Inception Score} of \texttt{2.27} and an \texttt{FID} of \texttt{0.12}.

\begin{figure}[htbp]
	\centering

	\begin{subfigure}[t]{0.48\linewidth}
		\centering
		\includegraphics[width=\linewidth]{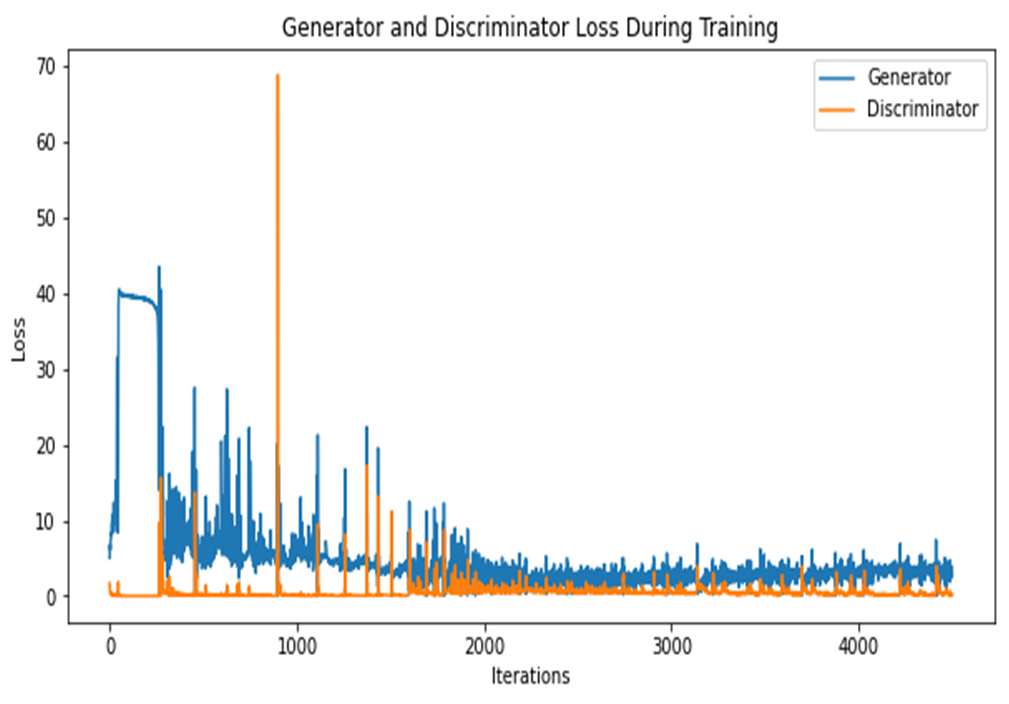}
		\caption{DCGAN training loss for the discriminator and generator on the BanglaLekha Isolated dataset.}
		\label{fig:loss}
	\end{subfigure}
	\hfill
	\begin{subfigure}[t]{0.48\linewidth}
		\centering
		\includegraphics[width=\linewidth]{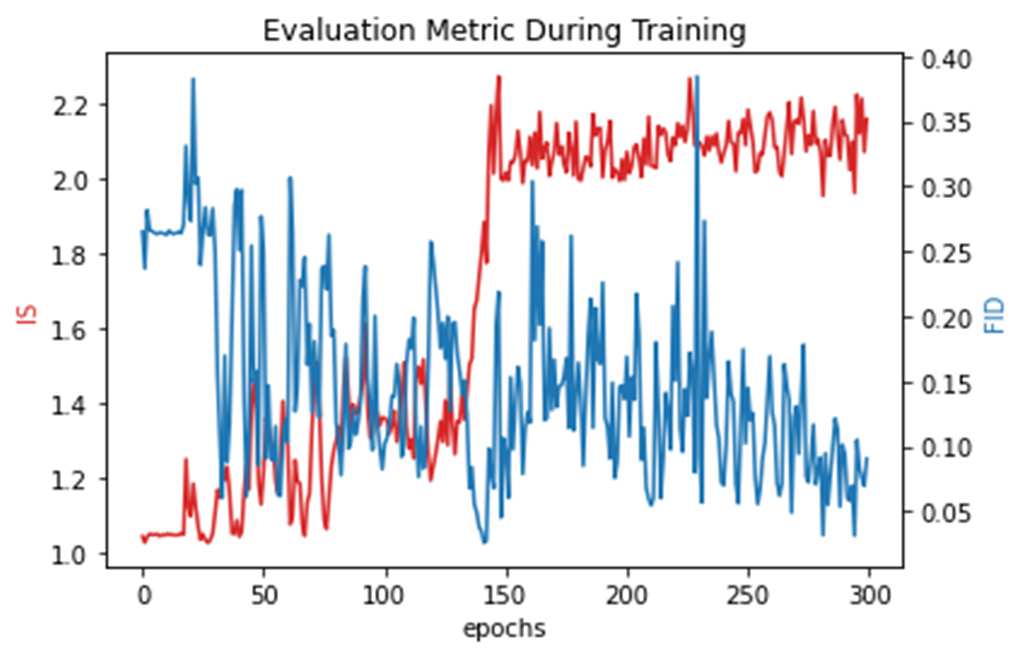}
		\caption{Evaluation metrics during DCGAN training.}
		\label{fig:eval_metric}
	\end{subfigure}

	\caption{Training behavior of the DCGAN model, showing generator--discriminator loss dynamics and evaluation metric trends during training.}
	\label{fig:training_curves}
\end{figure}

The low \texttt{FID} values demonstrate strong \textbf{\textit{distributional similarity}} between \textbf{\textit{real}} and \textbf{\textit{synthetic samples}}, while the moderate \texttt{IS} values point to some remaining limits in \textbf{\textit{diversity}}. Because both metrics depend on \texttt{ImageNet}-trained features, they do not fully capture \textbf{\textit{perceptual fidelity}} for \textbf{\textit{chest X-ray images}}, so their interpretation requires caution. For this reason, we place greater weight on \textbf{\textit{downstream classifier improvements}}, which provide a more reliable measure of \textbf{\textit{practical utility}}. Looking ahead, we plan to incorporate \textbf{\textit{domain-adapted metrics}} such as \texttt{Kernel Inception Distance} (\texttt{KID}) and the \texttt{Structural Similarity Index} (\texttt{SSIM}), which are better suited for evaluating \textbf{\textit{generative models}} in \textbf{\textit{non-natural image domains}}.

\begin{figure}[htbp]
	\centerline{\includegraphics[width=0.90\linewidth]{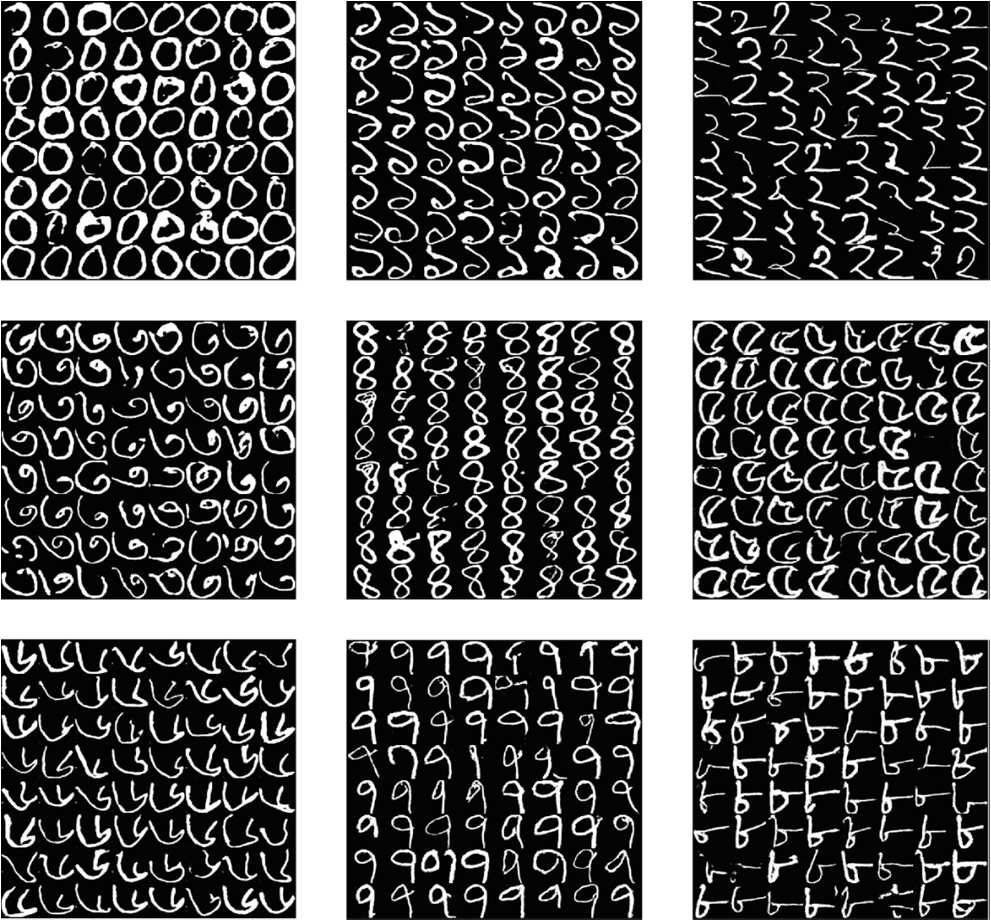}}
	\caption{DCGAN Real and Fake Score of Generated Image During Training on the Bangla Lekha Isolated dataset.}
	\label{fig:realfake_score}
\end{figure}

As the \textbf{\textit{generator}} and \textbf{\textit{discriminator}} losses approach stability and the difference between \texttt{actual} and \texttt{fake} scores steadily decreased. Figures \ref{fig:loss} to \ref{fig:eval_metric} show how training proceeded with steady \textbf{\textit{adversarial behavior}}. These patterns show that \texttt{spectral normalization} and \texttt{WGAN-GP} were used to stabilize training and prevent \textbf{\textit{mode collapse}}. This is proven by an \textbf{\textit{ablation study}}, where \texttt{spectrum normalization} lowers the \texttt{FID} score by roughly \texttt{12\%} and the \texttt{gradient penalty} improves it by an additional \texttt{9\%}.

\subsection{Embedding Visualization}

\begin{figure}[htbp]
	\centering

	\begin{subfigure}[t]{0.48\linewidth}
		\centering
		\includegraphics[width=\linewidth]{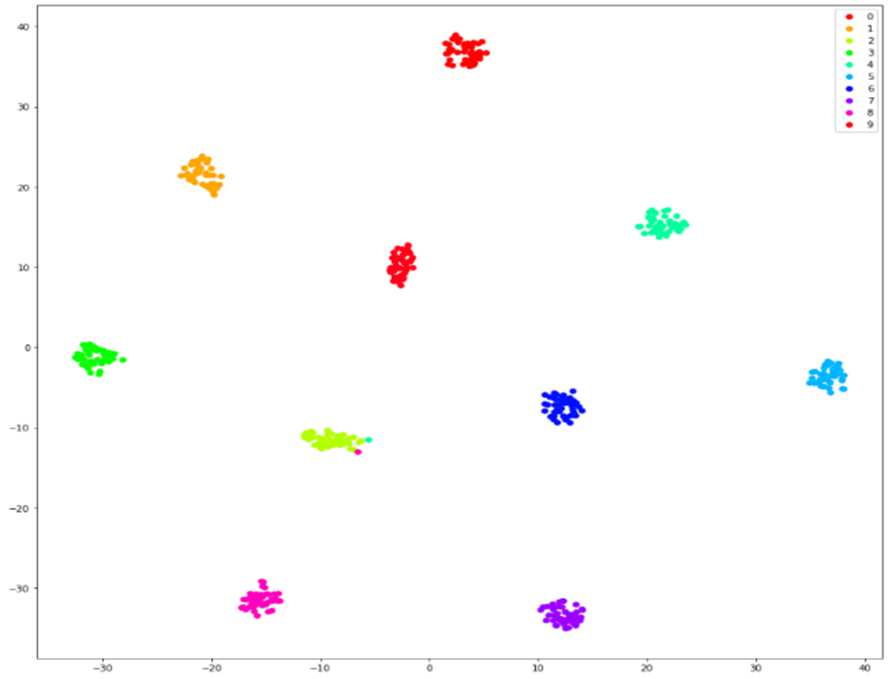}
		\caption{PCA 2-D plot of generated Bangla numeric characters.}
		\label{fig:pca}
	\end{subfigure}
	\hfill
	\begin{subfigure}[t]{0.48\linewidth}
		\centering
		\includegraphics[width=\linewidth]{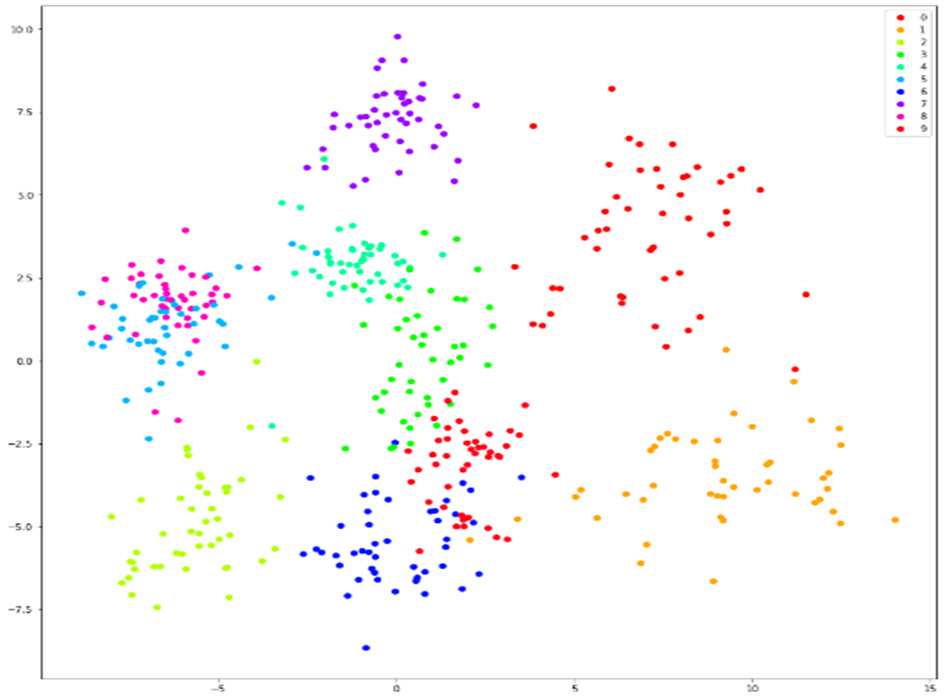}
		\caption{t-SNE 2-D plot of generated Bangla numeric characters.}
		\label{fig:tsne}
	\end{subfigure}

	\caption{Embedding-space visualization of real and GAN-generated Bangla numeric characters using two dimensionality reduction techniques. The PCA plot highlights the global variance structure and shows the overall overlap between real and synthetic samples, while the t-SNE plot provides a more detailed view of local neighborhood relationships, cluster formation, and class-wise mixing. Together, these visualizations indicate that the generated samples closely follow the underlying distribution of the real data while preserving class-specific structure and introducing useful stylistic diversity.}
	\label{fig:pca_tsne_combined}
\end{figure}

\textbf{\textit{Dimensionality reduction}} was used to figure out how well the \textbf{\textit{synthetic data}} resembled the true distribution. \textbf{\textit{Global variance}} was captured, as seen by the overlapping clusters of created and genuine \textbf{\textit{Bangla numerals}} in Figure \ref{fig:pca}'s \texttt{PCA} plot. Figure \ref{fig:tsne}'s \texttt{t-SNE} show, including synthetic samples into the actual class clusters, offers more detailed information. This implies that the approach added beneficial \textbf{\textit{stylistic diversity}} yet preserved \textbf{\textit{class-specific structure}}.

\subsection{Downstream Classifier Performance}

Three settings were used to assess \textbf{\textit{classifier performance}}: \textbf{\textit{actual data only}}, \textbf{\textit{real with traditional augmentation}}, and \textbf{\textit{real with GAN-augmented examples}}. Accuracy in \textbf{\textit{Bangla character recognition}} increased from \texttt{89.2\%} (\texttt{macro-F1 0.87}) to \texttt{91.8\%} (\texttt{0.90}) using conventional techniques and \texttt{94.7\%} (\texttt{0.94}) using \texttt{GAN} augmentation. Sensitivity at \texttt{90\%} specificity increased from \texttt{0.78} to \texttt{0.84} for \textbf{\textit{chest X-rays}}, while \texttt{AUROC} improved from \texttt{0.86} to \texttt{0.89} and then \texttt{0.93}. These results confirm that realistic samples generated by \textbf{\textit{GAN-based augmentation}} greatly improve classifiers in domains with \textbf{\textit{limited data}}. 

\section{Discussion}

\textbf{\textit{GAN-based augmentation}} outperformed \textbf{\textit{classical augmentation}} through generating appealing samples and enhanced \textbf{\textit{classifier performance}} in the \textbf{\textit{medical}} and \textbf{\textit{handwriting}} domains. Evaluations using \texttt{IS}, \texttt{FID}, and embeddings confirmed \textbf{\textit{fidelity}} and \textbf{\textit{variety}}, whereas stability techniques like \texttt{WGAN-GP} and \texttt{spectral normalization} prevented \textbf{\textit{mode collapse}}. However, \texttt{FID} relied on \texttt{ImageNet} features that are not entirely appropriate for \textbf{\textit{medical imaging tasks}}, and \texttt{IS} values remained low. This highlights that \textbf{\textit{downstream classifier increases}} are a more significant indicator of augmentation success, and intrinsic scores should be viewed with caution. Because synthetic data may still leak information, \textbf{\textit{privacy concerns}} are still present. 

For more dependable and secure deployment in delicate sectors, future research will investigate stronger generators like \texttt{StyleGAN} and \textbf{\textit{diffusion models}}, domain-specific evaluation methods, larger datasets, and \textbf{\textit{privacy-preserving training}}. To improve the technical depth and generalization of \textbf{\textit{generative augmentation}}, future research will build on this baseline by incorporating domain-specific fidelity metrics (\texttt{SSIM}, \texttt{LPIPS}) and higher-capacity generators such as \texttt{StyleGAN} and \textbf{\textit{diffusion models}}. 

\section{Conclusion}

This study shows that \textbf{\textit{GAN-based augmentation}} can meaningfully reduce \textbf{\textit{data scarcity}} in \textbf{\textit{handwriting}} and \textbf{\textit{medical imaging}}. On the \texttt{BanglaLekha-Isolated} dataset, accuracy increased from \texttt{89.2} to \texttt{94.7 percent} and \texttt{macro-F1} from \texttt{0.87} to \texttt{0.94}, while chest X-ray \texttt{AUROC} rose from \texttt{0.86} to \texttt{0.93} and sensitivity at \texttt{90 percent} specificity from \texttt{0.78} to \texttt{0.84}. Low \texttt{FID} scores (\texttt{0.24} for Bangla and \texttt{0.12} for X-rays) and embedding plots confirmed strong alignment between \textbf{\textit{real}} and \textbf{\textit{synthetic samples}}, and stability methods such as \texttt{WGAN-GP} and \texttt{spectral normalization} reduced \texttt{FID} by \texttt{9} and \texttt{12 percent}, preventing \textbf{\textit{mode collapse}}. 

Overall, the results show that a \textbf{\textit{unified GAN-based pipeline}} can reliably improve \textbf{\textit{classifier performance}} in \textbf{\textit{low-resource settings}}, though intrinsic metrics for medical images and privacy concerns remain important limitations. Future work will explore stronger generators such as \texttt{StyleGAN} and \textbf{\textit{diffusion models}}, adopt domain-specific metrics like \texttt{SSIM} and \texttt{LPIPS}, expand datasets, and apply \textbf{\textit{privacy-preserving strategies}} to support scalable and trustworthy \textbf{\textit{generative augmentation}}.

\bibliographystyle{plain}
\bibliography{references}

\end{document}